\pdfoutput=1

\documentclass[11pt]{article}

\usepackage{ACL2023}

\usepackage{times}
\usepackage{latexsym}
\usepackage{graphicx}
\usepackage[T1]{fontenc}

\usepackage[utf8]{inputenc}
\usepackage{amsmath}
\usepackage{graphicx}

\usepackage{microtype}

\usepackage{inconsolata}

\usepackage{multirow}
\usepackage{subcaption}
\usepackage{amsfonts,amssymb}
\usepackage{booktabs}

\title{PIP: Parse-Instructed Prefix for Syntactically \\ Controlled Paraphrase Generation}

\author{Yixin Wan \and Kuan-Hao Huang \and Kai-Wei Chang \\
  Computer Science Department, University of California, Los Angeles \\
  \texttt{elaine1wan@g.ucla.edu} \\ 
  \texttt{\{khhuang, kwchang\}@cs.ucla.edu}\\ }

\begin{document}
\maketitle
\begin{abstract}
Syntactically controlled paraphrase generation requires language models to generate paraphrases for sentences according to specific syntactic structures. 
Existing fine-tuning methods for this task are costly as all the parameters of the model need to be updated during the training process. 
Inspired by recent studies on parameter-efficient learning, we propose Parse-Instructed Prefix (PIP), a novel adaptation of prefix-tuning to tune large pre-trained language models on syntactically controlled paraphrase generation task in a low-data setting with significantly less training cost.
We introduce two methods to instruct a model's encoder prefix to capture syntax-related knowledge: direct initiation (PIP-Direct) and indirect optimization (PIP-Indirect).
In contrast to traditional fine-tuning methods for this task, PIP is a compute-efficient alternative with \(10 \times \) times less learnable parameters.
Compared to existing prefix-tuning methods, PIP excels at capturing syntax control information, achieving significantly higher performance at the same level of learnable parameter count.
\end{abstract}

\section{Introduction}
Syntactically controlled paraphrase generation (SCPG) has attracted increasing attention as it can diversify the generated paraphrases \cite{Iyyer2018AdversarialEG, Huang2021synpg, sun2021aesop}.
Given an input sentence and a target syntax specification, an SCPG model aims to generate paraphrases that satisfy the specific syntax requirement.
Such generation systems are promising in benefiting multiple application areas in natural language processing (NLP), such as text summarization \cite{fan-etal-2018-controllable}, dialogue systems \cite{niu-bansal-2018-polite,gao-etal-2020-paraphrase}, diverse question generation \cite{yu-jiang-2021-expanding}, creative generation \cite{tian-etal-2021-hypogen-hyperbole}, and improving the robustness of models \cite{Iyyer2018AdversarialEG, Huang2021synpg}.

However, prior studies on SCPG mainly explore fine-tuning strategies, which require updating the parameters of the entire language model to adapt to the newly included syntax information.
Therefore, many previously proposed methods suffer from tremendous training cost \cite{DBLP:journals/corr/abs-1910-13461,Raffel20t5,NEURIPS2020_1457c0d6}.
With the recent rise of larger pre-trained language models (PLMs), this problem has become even more imminent.
Nevertheless, a lightweight and more resource-efficient tuning method would allow easier application of large PLMs on the SCPG task.

Resource-efficient training methods such as prompt-tuning and prefix-tuning \cite{li-liang-2021-prefix,lester-etal-2021-power} have proven to be effective in tuning large PLMs on various NLP tasks, such as text classification \cite{xiao2021ptuningv2}, sequence labeling \cite{xiao2021ptuningv2}, and summarization \cite{li-liang-2021-prefix}.
Prefix-tuning freezes a PLM's parameters and optimizes a small task-oriented continuous prefix that is prepended to the model's Transformer layers.
It is a promising alternative to fine-tuning in a low-data setting with significantly fewer learnable parameters.
However, no previous literature has explored the potential of prefix-tuning on the SCPG task.

In light of the lack of previous studies, we are amongst the first to study the application of resource-efficient training methods on the SCPG task.
Our work has two main contributions.
To begin with, we are among the first to study prefix-tuning's application on the SCPG task as a compute-efficient alternative for fine-tuning.
Secondly, we propose \emph{parse-instructed prefix (PIP)}, a novel adaptation of prefix-tuning for enhanced syntax control in paraphrase generation.
Similar to prefix-tuning, PIP freezes all parameters of a PLM and only optimizes the prefix parameters, reducing the number of tune-able parameters to almost \(10 \times\) less than that required for fine-tuning.
Prefix-tuning methods initialize the prefix as continuous and completely free parameters.
For the SCPG task, this means that the prefix would need to learn the syntax control from scratch, since the PLMs were not pre-trained on any syntax-related task.
In contrast, PIP provides syntax-related guidance to the prefix, allowing for better capturing of syntax knowledge.
Specifically, we introduce two methods to guide the process of syntax knowledge capturing: direct initiation and indirect optimization.
We prove that prefix-tuning-based methods achieve promising performance in a low-data setting with significantly fewer learnable parameters.
In addition, our proposed PIP methods outperform prefix-tuning at the same level of training cost.\footnote{Our code for this work is available at \url{https://github.com/uclanlp/PIP}}

\section{Related Work}
\paragraph{Syntactically controlled paraphrase generation.} 
For the SCPG task, given a source sentence and a target syntax structure, a language model is trained to output a paraphrase sentence of the source sentence that (1) is semantically similar to the source sentence, and (2) conforms to the given target syntax structure, or the ``syntax control''.
Prior works mainly adopted encoder-decoder model structures and used sequence-to-sequence training for the SCPG task \cite{Iyyer2018AdversarialEG, kumar-etal-2020-syntax, Huang2021synpg, sun2021aesop}, while exploring different means to include syntax control signal during training.
The first type of approach encodes the source sentence and the target syntactic tree separately, then concatenates them at decoder input \cite{Iyyer2018AdversarialEG, kumar-etal-2020-syntax}.
The second type of approach concatenates linearized target constituency parse and source sentence at model input \cite{Huang2021synpg,Huang2022UnsupervisedSC,sun2021aesop}.
However, the aforementioned methods require updating all model parameters during tuning at a high training cost.

\paragraph{Prompt-tuning and prefix-tuning.}
Prompting \cite{NEURIPS2020_1457c0d6, Sun2020ConditionedNL} provides PLMs with a discrete task-specific ``prompt'' to generate task-related outputs without task-specific fine-tuning.
Prompt-tuning-based methods \cite{liu2021gpt, qin-eisner-2021-learning,lester-etal-2021-power,vu-etal-2022-spot,min2021noisy}, Prefix-Tuning \cite{li-liang-2021-prefix} and P-Tuning v2 \cite{xiao2021ptuningv2} derived from prompting and propose to only optimize a small sequence of continuous vectors.
However, since prefix-tuning learns a prefix that was initiated as a continuous vector with completely free parameters, the prefix would need to learn task information from scratch during training.
In addition, the training process for prefix-tuning does not allow for incorporation of any task-specific guidance.
In summary, existing prefix-based methods fail to consider both specific task instruction and model-learned knowledge.

\begin{figure}[t]
    \centering
    \includegraphics[width=7.2cm]{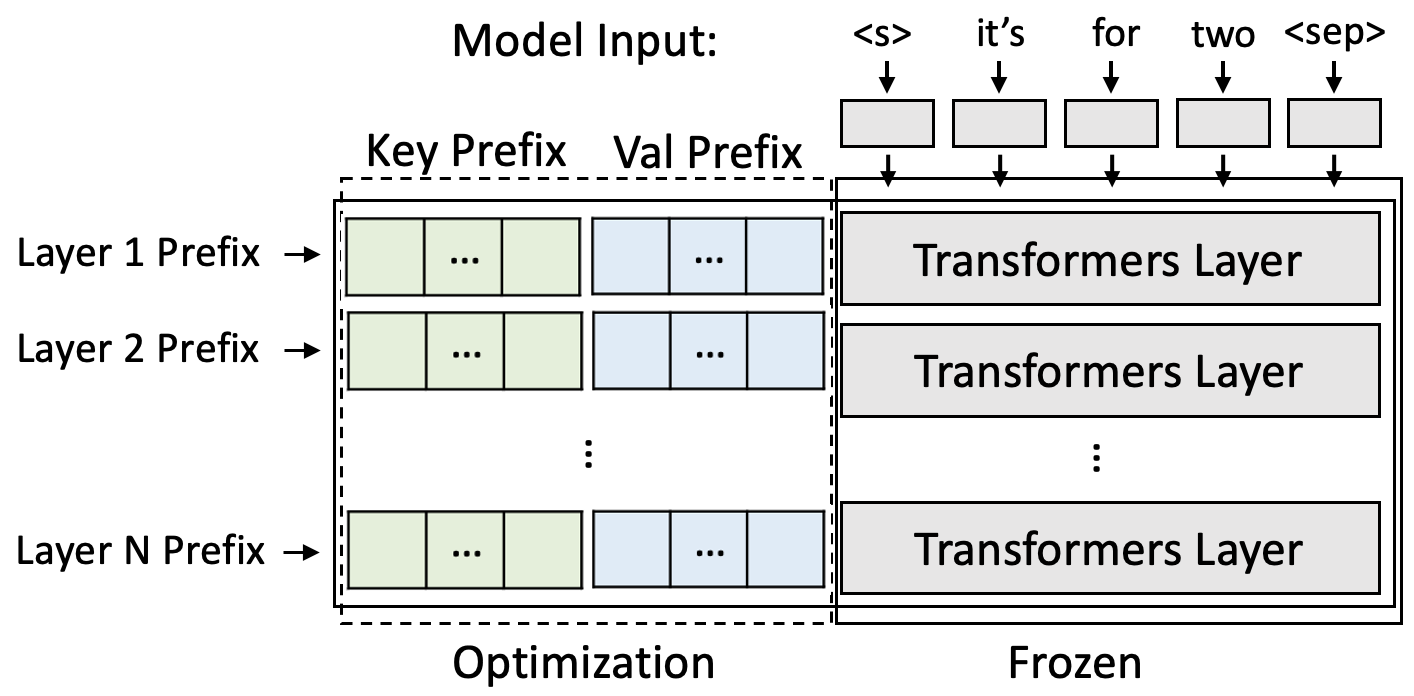}
    \caption{Structure of the prefix-tuning method.}
    \label{fig:prefix-tuning}
    \vspace{-1em}
\end{figure}

\begin{figure*}[t!]
\centering
\begin{subfigure}[b]{0.48\textwidth}
    \centering
    \includegraphics[width=\textwidth]{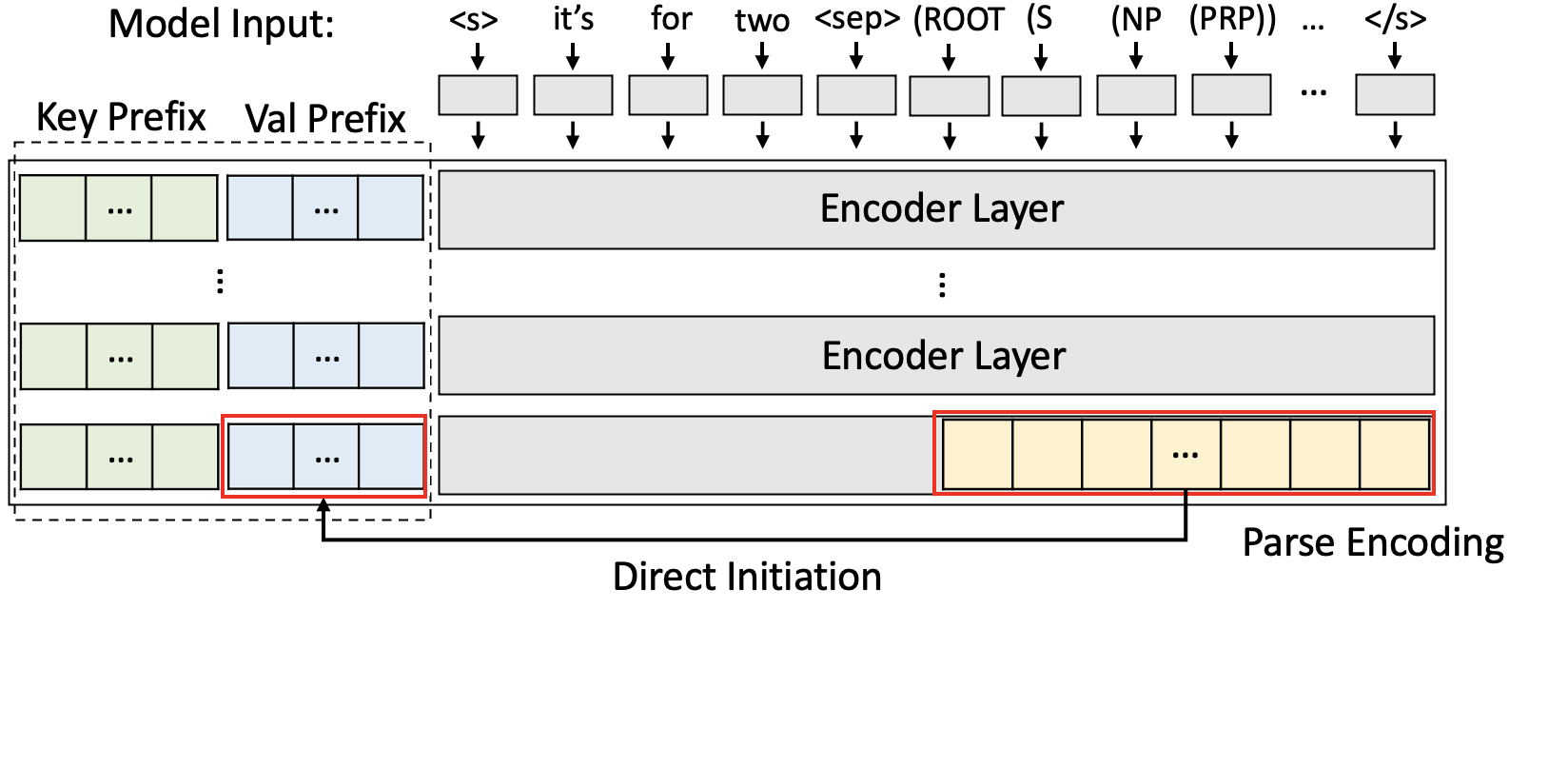}
    \caption{Direct Parse-Instructed Prefix (PIP-Direct).}
    \label{fig:model1}
\end{subfigure}
\begin{subfigure}[b]{0.48\textwidth}
    \centering
    \includegraphics[width=\textwidth]{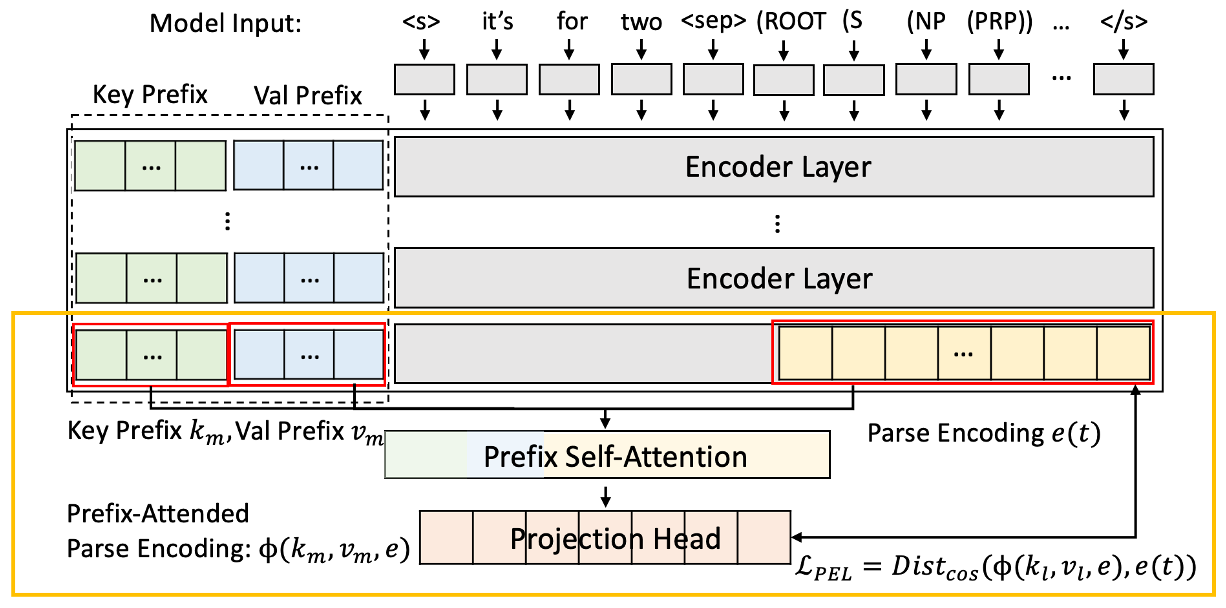}
    \caption{Indirect Parse-Instructed Prefix (PIP-Indirect).}
    \label{fig:model2}
\end{subfigure}
\caption{Structure of the proposed PIP-Direct and PIP-Indirect Models. Note that we only visualize the encoder of the BART model. The model decoder follows the regular prefix-tuning setting without modifications. In (a), the value prefix of the last encoder layer is directly initiated by the model encoding of the target parse. In (b), the Parse Encoding Loss (PEL) is calculated between the prefix-attended parse encoding and the model parse encoding.}
\label{fig:model}
\vspace{-1em}
\end{figure*}

\section{Method}
\paragraph{Problem formulation.}
Following previous studies \cite{Iyyer2018AdversarialEG, Huang2021synpg, Huang2022UnsupervisedSC}, we use an encoder-decoder model structure and utilize the constituency parse as the control signal.
Denote the source sentence as \(s_{src}\), the target parse as \(t\), and the target sentence as \(s_{tgt}\).
The goal of an SCPG model is to generate the target sentence \(s_{tgt}\) that is semantically similar to \(s_{src}\) while conforming to the syntax of \(t\).
In our study, the model is provided with \(s_{src}\) and \(t\) at input, and is supervised by the target sentence \(s_{tgt}\) at output during training.
Notice that previous methods \cite{sun2021aesop,Huang2022UnsupervisedSC} mainly fine-tune all PLM parameters and therefore suffer from high training costs.

\paragraph{Prefix-tuning.}
We investigate a resource-efficient method for training an SCPG model based on the prefix-tuning method \cite{li-liang-2021-prefix,xiao2021ptuningv2}.
\citet{li-liang-2021-prefix} freezes all pre-trained LM parameters and only optimizes a small sequence of continuous prefixes that are then prepended to keys and values of the attention module in the input layer of the model's encoder and decoder.
\citet{xiao2021ptuningv2} further extends this approach and applies prefixes to every layer of the model encoder and decoder.
We follow the previous approach \cite{li-liang-2021-prefix} and consider prepending additional prefix parameters to the key and value matrices of each Transformer layer in the PLM.
Specifically, we establish a prefix \(p\) with length \(\vert p \vert\) for a PLM with \(l\) layers and hidden dimension \(dim(h)\) and produce a set of key prefixes \(K_p = \{k_1, k_2, ..., k_l\}\) and a set of value prefixes \(V_p = \{v_1, v_2, ..., v_l\}\), 
where \(k_i, v_i \in \vert p \vert \times dim(h)\) denotes the key and value prefixes of layer \(i\), respectively.
For an encoder-decoder PLM, the key and value prefixes will then influence the model's encoding and decoding process through the attention mechanism, where the prefixes directly attend with the hidden states of the model.
Figure \ref{fig:prefix-tuning} visualizes structure of the prefix-tuning method.

\subsection{Parse-Instructed Prefix}
\paragraph{Intuition.}
In prefix-tuning, the learned prefix acts as a context that influences the encoding of inputs through extracting task-related information from the PLM \cite{li-liang-2021-prefix}.
However, as prefix-tuning optimizes a prefix with completely free parameters, the prefix is learned from scratch and is unable to receive task-specific guidance during tuning.
Since we use the constituency parse of target paraphrase as the control signal, which the PLM has never seen during pre-training, it will take a long time for the prefix to adapt to and learn the encoding for the control syntax. 
Specifically, for the SCPG task, 
the prefix will need to learn to:
(1) capture semantic and syntax information from model input, and
(2) combine the extracted semantic and syntax knowledge to produce an encoding for paraphrase generation under target syntax control.
Since the prefix is implemented as pre-pended parameters for keys and values in Transformer layers, it first retrieves semantic and syntax information by attending to the source sentence and target parse in model input.
Ideally, the prefix will then combine both the retrieved semantic and syntax information by influencing the output encoding.
The combined information will then be captured by the decoder to output a paraphrase that conforms to the syntactic control.
Therefore, guiding the prefix at encoder output to capture and leverage syntax knowledge will enhance the syntax control signal for improved performance on the SCPG task.


Therefore, we propose parse-instructed prefix (PIP) at the model's last encoder layer to ``instruct'' and augment the prefix's ability to capture task-specific syntax control for paraphrase generation.
Specifically, we introduce two PIP-based methods for better capturing of syntax control information: Direct Parse-Instructed Prefix (PIP-Direct) and Indirect Parse-Instructed Prefix (PIP-Indirect).
Different from prefix-tuning, where all prefix parameters are learned from scratch, PIP instructs the value prefix \(v_m\) of the last encoder layer with task-specific information.
Through the attention mechanism, the instructed value prefix will help to better capture syntax information in the model's encoding output.

\begin{table*}[htb]
\centering
\small
\begin{tabular}{lllcccccc}
\toprule
\textbf{Dataset} & \textbf{Model} & \textbf{\# Params} & \textbf{BLEU \(\uparrow\)} & \textbf{ROUGE-1\(\uparrow\)} & \textbf{ROUGE-2\(\uparrow\)} & \textbf{ROUGE-L\(\uparrow\)} & \textbf{TMA\(\uparrow\)} & \textbf{TED-3\(\downarrow\)}\\
\midrule
\multirow{4}*{\textbf{ParaNMT}} & Seq2seq & $139.42$M & $57.17$ & $79.20$ & $64.31$ & $78.81$ & $89.94$ & $0.59$ \\
\cmidrule{2-9}
& Prefix-Tuning & $15.83$M & $47.75$ & $73.85$ & $55.76$ & $73.17$ & $82.75$ & $1.12$\\
& PIP-Direct & $15.83$M & $49.61$ & $74.84$ & $57.39$ & $74.26$ & $85.81$ & $0.90$\\
& PIP-Indirect & $18.78$M & $\textbf{50.82}$ & $\textbf{75.58}$ & $\textbf{58.46}$ & $\textbf{75.00}$ & $\textbf{86.20}$ & $\textbf{0.87}$\\
\midrule
\multirow{4}*{\textbf{Pan}} & Seq2seq & $139.42$M & $42.68$ & $68.50$ & $47.59$ & $67.45$ & $76.61$ & $1.54$\\
\cmidrule{2-9}
& Prefix-Tuning & $15.83$M  & $38.11$ & $66.29$ & $43.66$ & $64.47$ 
 & $66.75$ & $2.41$ \\
& PIP-Direct & $15.83$M & $39.72$ & $67.29$ & $45.24$ & $65.73$ & $72.75$ & $\textbf{1.95}$\\
& PIP-Indirect & $18.78$M & $\textbf{40.44}$ & $\textbf{67.38}$ & $\textbf{45.48}$ & $\textbf{65.91}$ & $\textbf{72.95}$ & $1.98$\\
\midrule
\multirow{4}*{\textbf{MRPC}} & Seq2seq & $139.42$M & $48.17$ & $71.95$ & $53.06$ & $70.15$ & $87.24$ & $1.33$\\
\cmidrule{2-9}
& Prefix-Tuning & $15.83$M & $43.21$ & $69.63$ & $49.00$ & $66.90$ & $77.34$ & $2.42$\\
& PIP-Direct & $15.83$M & $45.00$ & $70.13$ & $50.66$ & $68.04$ & $\textbf{83.65}$ & $\textbf{1.88}$ \\
& PIP-Indirect & $18.78$M & $\textbf{45.33}$ & $\textbf{70.47}$ & $\textbf{50.71}$ & $\textbf{68.15}$ & $83.49$ & $1.89$ \\
\midrule
\multirow{4}*{\textbf{Quora}} & Seq2seq & $139.42$M & $49.93$ & $78.45$ & $58.29$ & $77.54$ & $79.64$ & $0.84$ \\
\cmidrule{2-9}
& Prefix-Tuning & $15.83$M & $42.4$ & $74.94$ & $51.21$ & $73.49$ & $70.69$ & $1.40$ \\
& PIP-Direct & $15.83$M & $\textbf{46.33}$ & $\textbf{77.20}$ & $\textbf{55.55}$ & $\textbf{76.00}$ & $76.00$ & $1.07$ \\
& PIP-Indirect & $18.78$M & $45.78$ & $76.68$ & $54.36$ & $75.39$ & $\textbf{77.34}$ & $\textbf{1.04}$ \\
\bottomrule
\end{tabular}
\caption{\label{results}
Experiment results. ``\# Params'' denotes the number of learnable parameters for each method. The PIP methods achieve highest performance amongst the three prefix-based methods on all valid and test datasets.
}
\vspace{-1em}
\end{table*}

\paragraph{Direct parse-instructed prefix.}
We propose Direct Parse-Instructed Prefix (PIP-Direct) as an intuitive way to enhance knowledge on syntax control at model encoding.
PIP-Direct directly updates the parameters of the value prefix at the last encoder layer with the model's encoding of the target parse.
That is, for an input with target syntax \(t\) and an LM encoder with \(m\) layers, we first retrieve the model's encoding output of the target parse, which we denote as \(e(t)\).
Then, for the set of model encoder's value prefixes \(V_p = \{v_1, v_2, ..., v_m\}\), we replace the value prefix of the last encoder layer with the parse encoding \(e(t)\).
The final value prefix prepended to the LM value state is then:
\vspace{-0.06cm}
\begin{equation*}
        v_i^{*} = \left\{\begin{array}{l}
        e(t),\; \text{if } i = m\\
        v_i,\; \text{otherwise} \\
        \end{array}
        \right.
    \end{equation*}
This method directly augments syntax-related information at the last model encoder layer, which enables the key prefix of the same layer to capture the augmented syntax knowledge through attention.
Structure of the direct initiation PIP is demonstrated on the left of Figure \ref{fig:model1}.

\paragraph{Indirect parse-instructed prefix.}
We propose Indirect Parse-Instructed Prefix (PIP-Indirect) as an alternative way to guide the capturing of target syntax knowledge at the last encoder layer.
Instead of directly replacing the prefix parameters, we utilize the Parse Encoding Loss (PEL) to indirectly augment the prefix's ability to capture syntax knowledge.
Given a parse input, the prefix will influence the original model encoding by attending to the parse input, resulting in a modified encoding of the parse information.
We can therefore augment the prefix's syntax information capturing by improving the ability to reconstruct the original parse encoding from the prefix-modified encoding.
For a target parse \(t\) with encoding \(e(t)\), we can obtain its prefix-modified encoding through an additional prefix self-attention layer \(\mathcal{A}(\cdot)\), in which the prefix directly attends to the parse encoding \(e(t)\). 
The prefix attention layer has the same structure as a prefix Transformer layer in the model, with the key prefix \(k_m\) and value prefix \(v_m\) of the last encoder layer \(m\) prepended to the attention key and value.
We denote the output encoding of this prefix self-attention layer as \(\mathcal{A}(k_m, v_m, e(t))\).
To examine the ability to reconstruct the original parse encoding \(e(t)\) from the prefix-modified encoding \(\mathcal{A}(k_m, v_m, e(t))\), 
we leverage a learnable projection head, denoted by \(\mathcal{H} (\cdot):dim(h) \rightarrow dim(h)\), to approximate the process of reconstruction.
We denote the output of the projection head as:
\(\phi (k_m, v_m, e(t)) = \mathcal{H}(\mathcal{A}(k_m, v_m, e(t)))\).
Then, we establish the PEL loss by measuring the projected output's cosine distance, or the reconstructed parse encoding, with the original model encoding of target parse, \(e(t)\).
The PEL loss is mathematically formulated as:
\vspace{-0.06cm}
\begin{equation*}
\begin{aligned}
    \mathcal{L}_{PEL} &= Dist_{cos}(\phi (k_m, v_m, e(t)), e(t)),
\end{aligned}
\normalsize
\end{equation*}
where \(Dist_{cos}\) denotes the cosine distance.
By integrating the PEL loss with the LM output loss during optimization, we can indirectly guide the prefix to better capture syntax-related information in model encoding.
The structure of PIP-Indirect is demonstrated on the right of Figure \ref{fig:model2}.


\section{Experiments}
We conducted experiments on our proposed PIP-Direct and PIP-Indirect methods, as well as two baseline methods for comparison.
All four training methods are implemented on the BART-base model \cite{DBLP:journals/corr/abs-1910-13461}.
For all models, we concatenate the source sentence and target parse as input, and train the models to output a paraphrase that follows the given target syntax.

\paragraph{Dataset.} 
We use ParaNMT \cite{chen-etal-2019-multi} as the training and validation dataset for all models.
Specifically, we sample 30,000 and 6,400 data entries from ParaNMT as our training set and dev set, respectively.
To test the models' abilities to generate paraphrases with syntax control in unseen domains, we follow previous work \cite{Huang2021synpg,Huang2022UnsupervisedSC} and apply the trained models on three mainstream test datasets: Quora 
\cite{iyer2017first}, MRPC \cite{dolan-etal-2004-unsupervised}, and PAN \cite{madnani-etal-2012-examining}.

\paragraph{Evaluation metrics.}
Conforming to prior works \cite{Huang2022UnsupervisedSC,sun2021aesop}, we evaluate generations of models on both alignment-based and syntactic conformation metrics.
Alignment-based metrics measure the similarity between target paraphrases and model-generated paraphrases.
We consider $4$ alignment-based evaluation metrics: BLEU \cite{Papineni2002BleuAM}, ROUGE-1, ROUGE-2, and ROUGE-L \cite{lin-2004-rouge}.
Syntactic conformation metrics measure the quality of syntactic control in generated paraphrases.
We consider $2$ syntactic conformation evaluation metrics: Template Matching accuracy (TMA) and Tree-Edit Distances score \cite{zhang1989simple} at height three (TED-3).

\paragraph{Baselines.}
We establish $2$ baseline training methods for our study.
The first baseline is the vanilla fine-tuning method that updates all parameters in the PLM, which we denote as \textbf{Seq2Seq},
In addition, we consider \textbf{prefix-tuning}, which freezes the PLM and learns a small set of parameters as the prefix, as the second baseline. 
Experiments on prefix-tuning in this study are based on our implementation of \citeauthor{li-liang-2021-prefix} and \citeauthor{xiao2021ptuningv2}'s work.

\paragraph{Implementation details.}
We train all baselines and our proposed PIP methods for $10$ epochs with batch size $64$.
At decoding stage, we use beam search with beam size $4$.
We use the AdamW optimizer \cite{Loshchilov2017DecoupledWD} and apply gradient clipping for training all models.
For fine-tuning, we set learning rate to $10^{-5}$ with a linear scheduler.
For prefix-tuning and PIP methods, we set learning rate to $3\times 10^{-4}$.


\paragraph{Results.}
Results of experiments show that our proposed PIP methods outperform prefix-tuning on the validation dataset and all test datasets by a significant margin in a low-data setting and at the same level of training cost.
Note that we separate results of the ``Seq2Seq'' model just for reference. 
We observe that PIP-Indirect achieves highest performance across all metrics among the $3$ prefix-tuning-based approaches on the validation set of ParaNMT, as well as on testsets Pan and MRPC.
PIP-Direct outperforms other prefix-tuning-based methods for the Quora dataset.


\paragraph{Analysis.}
We conduct additional ablation experiments to further validate experimental results.
Specifically, we examine if PIP-Indirect's performance gain is due to the effectiveness of design or the slightly higher parameter count compared to prefix-tuning.
We experiment on prefix-tuning with an additional linear layer during prefix construction, denoted as Prefix-Tuning-Large.
Prefix-Tuning-Large has $31.56$M learnable parameters, $12.78$M more than the PIP-Indirect method. 

\begin{table}[t]
\centering
\small
\begin{tabular}{llccc}
\toprule
\textbf{Model} & \textbf{BLEU \(\uparrow\)} & \textbf{TMA\(\uparrow\)} & \textbf{TED-3\(\downarrow\)}\\
\midrule
PIP-Direct & $49.61$ & $85.81$ & $0.90$  \\
PIP-Indirect & $\textbf{50.82}$ & $\textbf{86.20}$ & $\textbf{0.87}$  \\
Prefix-Tuning & $47.75$ & $82.75$ & $1.12$  \\
Prefix-Tuning-Large & $47.67$ & $82.94$ & $1.10$  \\
\bottomrule
\end{tabular}
\caption{\label{results2}
Ablation experiment results.
}
\vspace{-1em}
\end{table}

Table \ref{results2} demonstrates results of the ablation experiment on ParaNMT’s validation dataset.
We observe that although having more parameters, Prefix-Tuning-Large fails to outperform the PIP methods.
In addition, Prefix-Tuning-Large even fails to outperform the original Prefix-Tuning method, which only has $15.83$M parameters.
This provides further insights that 1) a larger number of parameters in prefix-tuning-based methods does not guarantee performance gain on downstream tasks, and 2) outstanding performance of the PIP methods on SCPG task is due to the effectiveness of method design.

\section{Conclusion}
This research is amongst the first to study resource-efficient training methods for syntactically controlled paraphrase generation tasks.
In this work, we proposed Parse-Instructed Prefix (PIP), a compute-efficient method that only requires \(10 \times\) less learnable parameters than traditional fine-tuning methods.
We introduce Direct and Indirect PIP methods to further improve prefix-tuning's performance by providing task-specific guidance and augmenting task-related knowledge at the fine-tuning stage. 
Through extensive experiments, we find out that both PIP-Direct and PIP-Indirect outperform prefix-tuning in a low-data setting at the same level of parameter count, and are promising as a resource-efficient alternative to fine-tuning.
With ablation studies, we further validate that the performance gain of the proposed PIP methods is due to the effectiveness of the design.

\section*{Acknowledgments}
We thank anonymous reviewers for their helpful feedback. We thank the UCLA-NLP group for their valuable discussions and comments. This work is supported in part by National Science Foundation (NSF) via Award No.~2200274 and a CISCO research grant. 

\section*{Limitations}
We identify some limitations of our study.
First, due to a lack of computing resources, we were not able to experiment with even larger pre-trained language models such as GPT-2 and GPT-3.
In future explorations, we would like to seek the opportunity to investigate the potential of instructed prefix-tuning on even larger-scale language models across a variety of generation tasks.
Second, since our work are amongst the first to explore the application of prefix-tuning on the task of syntactically-controlled paraphrase generation, we were not able to identify state-of-the-art prior works on the same subject in the field to establish comparison with.
We believe, however, that with our demonstration of promising application of prefix-tuning for SCPG, researchers will soon propose new ideas to utilize prefix for tuning large PLMs on this task at even lower training costs.

\section*{Ethics Considerations}
The experimented and proposed model in this study is based on large-scale Pre-trained Language Models (PLM).
Recent studies have revealed that large PLMs that are trained on large textual corpora might learn or even amplify the bias existing in its training dataset.
Therefore, since our method is established on top of a large PLM that is potentially at risk of demonstrating or amplifying bias, we recommend that potential harm and biases be evaluated before deploying our method in real-world situations.











\end{document}